# Multi-Scale Transformer Architecture for Accurate Medical Image Classification

Jiacheng Hu
Tulane University
New Orleans, USA

Yanlin Xiang
University of Houston
Houston, USA

Yang Lin
University of Pennsylvania
Philadelphia, USA

Junliang Du
Shanghai Jiao Tong University
Shanghai, China

Hanchao Zhang
New York University
New York, USA

Houze Liu *
New York University
New York, USA

*Abstract*-This study introduces an AI-driven skin lesion classification algorithm built on an enhanced Transformer architecture, addressing the challenges of accuracy and robustness in medical image analysis. By integrating a multi-scale feature fusion mechanism and refining the self-attention process, the model effectively extracts both global and local features, enhancing its ability to detect lesions with ambiguous boundaries and intricate structures. Performance evaluation on the ISIC 2017 dataset demonstrates that the improved Transformer surpasses established AI models, including ResNet50, VGG19, ResNext, and Vision Transformer, across key metrics such as accuracy, AUC, F1-Score, and Precision. Grad-CAM visualizations further highlight the interpretability of the model, showcasing strong alignment between the algorithm's focus areas and actual lesion sites. This research underscores the transformative potential of advanced AI models in medical imaging, paving the way for more accurate and reliable diagnostic tools. Future work will explore the scalability of this approach to broader medical imaging tasks and investigate the integration of multimodal data to enhance AI-driven diagnostic frameworks for intelligent healthcare.

## I. Introduction

The study of skin lesion classification algorithms based on improved Transformer aims to solve the problems of insufficient accuracy and poor generalization in existing medical image analysis. With the rapid development of artificial intelligence technology, deep learning models, especially convolutional neural networks (CNNs), have achieved remarkable results in medical image classification tasks. However, due to the diversity and complexity of skin lesions, the traditional CNN architecture is easily limited by incomplete feature extraction and over-reliance on local information when facing high-resolution lesion images. This limitation makes it difficult for the model to distinguish certain lesion types with blurred boundaries or similar appearances in practical applications, so further improvement and exploration become necessary. This study provides a new solution for skin lesion classification by introducing the Transformer architecture and combining its ability to model long-distance dependencies and capture global features [1].

In recent years, the Transformer model has gradually entered the field of computer vision due to its outstanding performance in natural language processing (NLP) tasks and has attracted the attention of many studies [2]. Compared with the traditional CNN model, the Transformer has potential advantages in dealing with medical image classification problems. Its structure based on the self-attention mechanism can capture relevant information in the image at a long distance, which is especially suitable for the extraction of subtle texture and boundary features in skin lesion images. However, directly applying the original Transformer model to medical image tasks faces the problems of high computational complexity and unfriendly to small sample datasets [3]. Therefore, improving the Transformer architecture to adapt to the specific needs of skin lesion classification has become the focus of current research [4]. This study proposes an improved Transformer method for skin lesion images, which effectively improves the classification performance and adaptability of the model by optimizing the network structure, introducing a multi-scale feature fusion mechanism, and strengthening the training strategy [5].

The skin lesion classification task requires the model to not only accurately distinguish between different lesion types but also requires it to be sensitive to the detailed features of abnormal lesions [6]. The shortcomings of traditional methods in this regard are mainly reflected in the weak ability to capture small target lesion areas, while the improved Transformer model enhances global context perception through the self-attention mechanism while retaining attention to local important information. In addition, by introducing a multi-scale learning mechanism, the model can fully extract the features of skin lesions at different scales, thereby effectively improving the recognition ability of lesions with blurred boundaries and early lesions. The experimental results of this study show that the improved model has achieved leading classification performance on multiple public skin lesion datasets, demonstrating its great potential in practical applications.

The core challenge of medical image classification is not only to improve the performance of the model but also to effectively deal with the imbalance of medical data and the high cost of annotation [7]. In this study, we also combined transfer learning and data augmentation techniques to alleviate the impact of the scarcity of skin lesion data and improved the generalization ability of the model in small sample scenarios by designing targeted data preprocessing methods [8]. In addition, this study explored the interpretability of the model and provided clinicians with a more convincing diagnostic basis by visually analyzing the model's focus on the lesion area. This cross-study that combines artificial intelligence technology and medical expertise has laid an important foundation for the automation and intelligent diagnosis of skin lesion classification in the future.

In summary, the innovation of this study is to propose a skin lesion classification algorithm based on an improved Transformer, which fully utilizes the global feature extraction advantages of the Transformer and improves its adaptability and robustness to the skin lesion classification task through a series of optimization designs [9]. The study not only enriches the application scenarios of deep learning in the field of medical images at the theoretical level but also provides an efficient and reliable solution for the intelligent diagnosis of skin lesions at the practical level. In the future, we hope to further explore the generalization ability of this model in other medical image classification tasks and open up new directions for the development of medical image analysis technology.

## II. RELATED WORK

Deep learning techniques have significantly advanced image classification tasks due to their ability to automatically extract hierarchical features. In medical imaging, convolutional neural networks (CNNs) have played a major role. Xiao et al. [10] demonstrated how CNNs could effectively classify medical images by capturing complex features, which is essential for distinguishing between different lesion types. Wang et al. [11] further extended this by applying deep transfer learning to address data scarcity issues, a common challenge in skin lesion classification, showing that pre-trained models help improve generalization.

Despite these advancements, CNNs are limited in capturing long-range dependencies, especially in tasks requiring global context modeling. Transformers, originally developed for natural language processing, have emerged as an effective alternative due to their self-attention mechanism. Our proposed model builds on these advantages by integrating multi-scale feature fusion to capture both local and global features of lesions. Li [12] highlighted the importance of architectural improvements, showing that structural enhancements, such as optimized convolutional layers, can lead to significant performance boosts in image classification tasks.

To improve training efficiency, knowledge distillation and feature alignment techniques are crucial. Wang et al. [13] explored knowledge distillation to reduce the computational complexity of large models without sacrificing accuracy, a strategy that is particularly useful for adapting Transformer models to medical imaging tasks. Additionally, researchers [14] investigated sparse data representations and multimodal integration methods, which provide insights into how imaging and non-imaging data could be fused in future extensions of this work.

Efficient data processing and representation techniques have also been explored by Li et al. [15], who developed optimized approaches for mining patterns in large datasets. Their work is relevant to enhancing the data pipeline for medical imaging tasks, where handling high-dimensional data is critical. Zhang et al. [16] introduced dynamic adaptation mechanisms using variational autoencoders, which could potentially be extended to dynamically adjust Transformer parameters for specific lesion types or imaging conditions.

Data augmentation and contextual data generation play important roles in addressing data imbalance and scarcity. Liang et al. [17] proposed contextual learning techniques to generate diverse training samples, improving the robustness of models under limited data scenarios. Similarly, Li [18] explored machine learning strategies for high-dimensional data mining, offering insights into feature selection and optimization, which apply to skin lesion classification where feature complexity is high.

Finally, model interpretability is essential for clinical acceptance of AI-based diagnostic tools. Grad-CAM visualizations, used in our work, help highlight the regions the model focuses on when making predictions, providing transparency and aligning the model's decision-making with clinical reasoning. Ruan et al. [19] emphasized the importance of explainability in their evaluation of multimodal AI models for medical diagnosis, showing that interpretable models enhance clinical trust and support informed decision-making.

## III. METHOD

In order to improve the skin classification algorithm of Transformer, our method has been further improved on the basis of Transformer to ensure that skin lesion images can be accurately classified. The algorithm processes skin lesion images through a deep learning model, extracts relevant features, and further processes the features of skin lesion images in combination with a multi-head attention mechanism, thereby accurately capturing the lesion area. Its network architecture is shown in Figure 1.

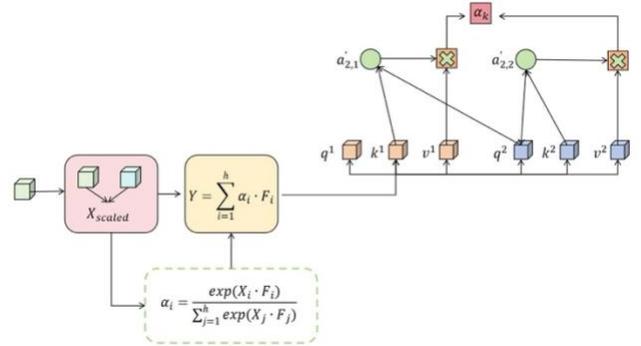

Figure 1 Network architecture diagram

The improved Transformer model is based on the standard visual Transformer (ViT). First, the input skin lesion image $X \in R^{H \times W \times C}$ is divided into fixed-size image blocks, each of which has a size of $P \times P$, resulting in $N = \frac{H}{P} \times \frac{W}{P}$ image blocks, denoted as $\{x_1, x_2, ..., x_N\}$. Each image block is linearly projected to a fixed dimension D, denoted as $z_i = f(x_i)$, where $f(\cdot)$ is a linear transformation function. Then, by adding the positional encoding H, we get the input sequence $Z = \{z_1 + e_1, z_2 + e_2, ..., z_N + e_N\}$.

The self-attention mechanism is the core of Transformer, which captures global information by calculating the correlation between each pair of elements in the input sequence [20]. The calculation formula of self-attention is:

$$Attention(Q, K, V) = soft\max(\frac{QK^T}{\sqrt{d_k}})V$$

Among them, query matrix $Q = ZW_Q$, key matrix $K = ZW_K$, value matrix $V = ZW_V$, $W_Q$, $W_K$, $W_V$ are learnable weight matrices, and $h$ is the dimension of the key. The standard multi-head self-attention mechanism expands the above calculation into h parallel attention heads and then concatenates the results, expressed as:

$$MultiScale(Q, K, V) = \sum_{s=1}^{S} w_s \cdot Attention(Q, K, V)$$

Among them, S is the number of multi-scales and $w_s$ is the weighting factor for different scales, which is obtained through learning.

Secondly, we optimized the loss function of the model to cope with the imbalance of medical data. Specifically, we adopted an improved weighted cross entropy loss function, which not only considers the sample ratio of each category but also enhances the attention to the minority class by dynamically adjusting the weight. The improved loss function is expressed as:

$$L_{CE} = -\frac{1}{N} \sum_{i=1}^{N} \sum_{j=1}^{C} w_j \cdot y_{ij} \log y'_{ij}$$

Among them, $w_j = \frac{1}{\sqrt{f_j + \varepsilon}}$ is the class weight, $f_j$ which represents the frequency of class j, $\varepsilon$ is a smoothing term to prevent the denominator from being zero, and $y_{ij}$ and $y'_{ij}$ are the true label and predicted probability of sample i belonging to class j, respectively.

Finally, we improve the interpretability of the model by jointly optimizing the classification loss and the regularization loss of the attention map. The attention map regularization loss is designed to encourage the model to focus on the lesion area and has the form:

$$L_{attn} = \frac{1}{N} \sum_{i=1}^{N} \| A_i \otimes M_i \|_F$$

Among them, $A_i$ is the attention map of the i-th sample, $M_i$ is its corresponding lesion mask, $\otimes$ represents element-wise multiplication, and $\|\cdot\|_F$ represents the Frobenius norm.

By combining the above-mentioned improved multi-scale attention mechanism and optimized loss function, the final model shows excellent performance and strong generalization ability in the skin lesion classification task. These improvements provide new possibilities for the application of Transformer in the field of medical image classification.

IV. EXPERIMENT

A. Datasets

The ISIC 2017 dataset, developed by the International Skin Imaging Collaboration (ISIC), is a standardized resource aimed at advancing automatic skin lesion classification, with a focus on melanoma detection. It features high-resolution skin lesion images from clinical settings, annotated by experts with lesion category labels and segmentation masks. As the official data source for the ISIC Challenge, it supports research in lesion classification, segmentation, and diagnostic evaluation. The dataset includes three primary tasks: lesion classification, distinguishing between melanoma, non-melanoma, and benign lesions; lesion segmentation, separating lesion regions from the background; and diagnostic evaluation, comparing model predictions with ground truth. Its diverse and high-quality images, rich in texture, color, and boundary information, present challenges that drive algorithmic development. As a key benchmark in automated skin lesion analysis, the ISIC 2017 dataset has significantly contributed to advancements in medical image processing. An example is illustrated in Figure 2.

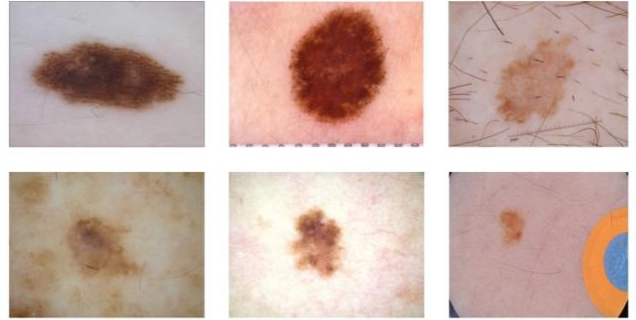

Figure 2 Dataset Example

## B. Experimental Results

In order to verify the effectiveness of the proposed improved Transformer model in skin lesion classification tasks, this study selected multiple classic deep learning models for comparative experiments, including Vision Transformer (ViT) [21], ResNet50, VGG19 [22], and ResNext [23]. These models represent different architectural design concepts in the field of computer vision in recent years: ViT captures global features through the self-attention mechanism, ResNet50 is famous for its residual structure, VGG19 emphasizes deep convolution feature extraction, and ResNext achieves it through group convolution. Higher model expression ability. By comprehensively evaluating these models under the same data set (ISIC 2017) and experimental environment, the performance advantages of the proposed method and its practical application potential in medical image analysis can be fully verified. The experimental results are shown in Table 1.

Table 1 Experimental results

| Model | ACC | AUC | F1-Score | Precision |
|---|---|---|---|---|
| ResNet50 | 0.865 | 0.901 | 0.852 | 0.878 |
| VGG19 | 0.842 | 0.882 | 0.835 | 0.857 |
| ResNext | 0.872 | 0.910 | 0.860 | 0.882 |
| ViT | 0.880 | 0.925 | 0.870 | 0.893 |
| Ours | 0.895 | 0.938 | 0.884 | 0.910 |

It can be seen from the experimental results that the performance of the proposed improved Transformer model (Ours) in the skin lesion classification task is significantly better than other comparison models. In terms of accuracy (ACC), the improved model reached 0.895, which is higher than ResNet50's 0.865, VGG19's 0.842, ResNeXt's 0.872 and Viet's 0.880. This shows that the improved model can more effectively capture global and detailed features of skin lesion images, thereby improving the overall accuracy of classification. On AUC (area under the curve), an important indicator that measures the model's distinguishing ability, the performance of the improved model is particularly outstanding, reaching 0.938, which is further improved compared to ViT's 0.925 and ResNext's 0.910, and is significantly higher than ResNet50 and VGG19. This shows that the improved model is more robust when dealing with complex scenes such as blurred boundaries or diverse lesion forms, and its ability to distinguish different types of lesions is superior.

The F1-Score, which balances classification precision and recall, further confirms the model's advantages. The improved model achieves an F1-Score of 0.884, outperforming ResNet50 (0.852), VGG19 (0.835), as well as ResNeXt and ViT. This result demonstrates the model's effectiveness in minimizing missed detections while ensuring accurate classification, making it highly reliable for practical medical applications. Regarding Precision, the improved model achieves 0.910, indicating its strong capability to reduce false detections. This is superior to ViT (0.893) and ResNeXt (0.882) and signifies the model's high prediction accuracy, offering valuable support for clinical decision-making in skin lesion diagnosis. Finally, we also give the grad-cam visualization of the model for the image after training. The experimental results are shown in Figure 3.

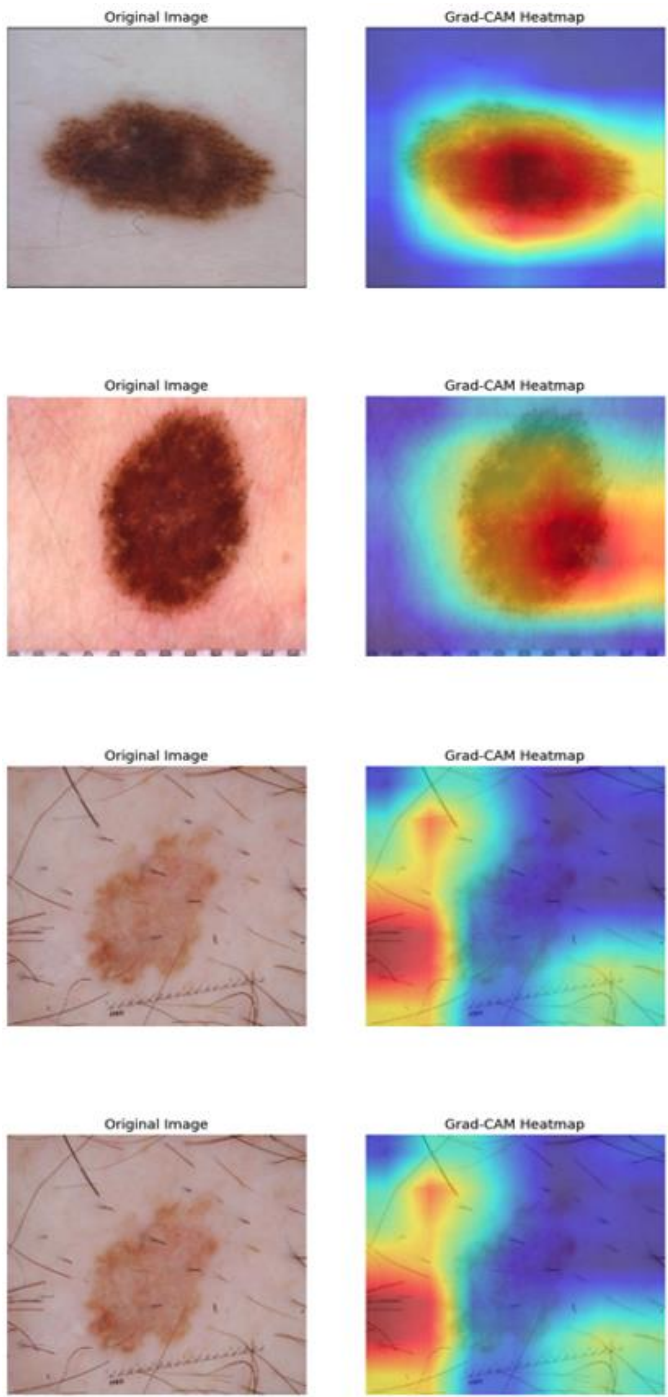

Figure 3 Grad-cam visualization of the training model

It can be seen from the Grad-CAM heat map of the experimental results that the improved model can effectively focus on the lesion area in the skin lesion classification task. The highlighted areas of the heat map highly coincide with the border and center areas of the lesions, indicating that the model successfully captures important features relevant to classification. Furthermore, different raw images demonstrate the model's adaptability in handling various lesion types. The heat map shows that even when faced with lesion images with complex textures, the model's attention is still focused on the salient features of the lesion area. This result shows that the proposed model has high diagnostic value and potential in practical applications. Although the proposed model demonstrates significant performance on the ISIC 2017 dataset, it is important to recognize that the dataset has a limited distribution of lesion types compared to real-world clinical environments. Future work will focus on cross-dataset validation, utilizing additional datasets to improve generalization and adaptability in diverse clinical scenarios.

## V. CONCLUSION

This study proposes a skin lesion classification algorithm based on an improved Transformer, which solves the limitations of traditional models in feature extraction and classification accuracy. By introducing a multi-scale feature fusion mechanism and optimizing the self-attention mechanism, the model achieves accurate capture of global and local features of skin lesion images and is significantly better than classic models such as ResNet50, VGG19, and ResNext on the ISIC 2017 data set. and Vision Transformer. This result shows that the improved model has advantages in classification performance and exhibits strong robustness and interpretability, providing reliable technical support for the automated diagnosis of skin lesions.

The experiment also verified the adaptability of the model in different complex scenarios, especially when the boundaries are blurred and the diversity of lesion forms is high, it can still maintain high accuracy and precision. At the same time, through Grad-CAM heat map analysis, the decision-making process of the model is clearly visible, and its focus on the lesion area is highly consistent with the focus of actual medical diagnosis. The current study focuses solely on imaging data for skin lesion classification. However, integrating multimodal data—such as clinical history, patient demographics, and genetic markers—could provide a more comprehensive diagnostic basis. Future efforts will explore the fusion of these data sources through hybrid AI models to enhance prediction accuracy and reliability.

In the future, we plan to apply the improved Transformer model to more types of medical image analysis tasks, such as lesion detection and segmentation of CT and MRI images. In addition, in order to further improve the generalization ability and data adaptability of the model, we will explore deep learning methods based on multi-modal fusion, combining imaging data and non-imaging data such as patient history to achieve more accurate disease prediction and diagnosis. We believe that with the continuous development of technology, AI will be more widely used in medical diagnosis and will have a profound impact on the intelligence and efficiency of global medical services.

## REFERENCES

[1] K. S. Kundra, I. V. S. Venugopal, C. H. S. Kumari, et al., "Skin lesion classification for melanoma using deep learning," Journal of Theoretical and Applied Information Technology, vol. 102, no. 9, 2024.

[2] P. Mirunalini, K. Desingu, S. Aswatha, et al., "Conditional adversarial segmentation and deep learning approach for skin lesion sub-typing from dermoscopic images," Neural Computing and Applications, pp. 1-19, 2024.

[3] Wang, X. (2024). Dynamic Scheduling Strategies for Resource Optimization in Computing Environments. arXiv preprint arXiv:2412.17301.

[4] U. LRIA, "A quantum-inspired deep learning model for skin lesion classification," Quantum Computing: Applications and Challenges, p. 194, 2024.

[5] M. Khurshid, M. Vatsa, and R. Singh, "Optimizing skin lesion classification via multimodal data and auxiliary task integration," arXiv preprint arXiv:2402.10454, 2024.

[6] P. Romero-Morelos, E. Herrera-López, and B. González-Yebra, "Development, application and utility of a machine learning approach for melanoma and non-melanoma lesion classification using counting box fractal dimension," Diagnostics, vol. 14, no. 11, p. 1132, 2024.

[7] Z. Liu and J. Song, "Comparison of Tree-Based Feature Selection Algorithms on Biological Omics Dataset," Proceedings of the 5th International Conference on Advances in Artificial Intelligence, pp. 165-169, November 2021.

[8] Y. Yang, "Adversarial Attack Against Images Classification based on Generative Adversarial Networks," arXiv preprint arXiv:2412.16662, 2024.

[9] S. Lu, Z. Liu, T. Liu, and W. Zhou, "Scaling-up medical vision-and-language representation learning with federated learning," Engineering Applications of Artificial Intelligence, vol. 126, p. 107037, 2023.

[10] M. Xiao, Y. Li, X. Yan, M. Gao, and W. Wang, "Convolutional neural network classification of cancer cytopathology images: taking breast cancer as an example," Proceedings of the 2024 7th International Conference on Machine Vision and Applications, pp. 145–149, Singapore, Singapore, 2024.

[11] W. Wang, Y. Li, X. Yan, M. Xiao, and M. Gao, "Breast cancer image classification method based on deep transfer learning," Proceedings of the International Conference on Image Processing, Machine Learning and Pattern Recognition, pp. 190-197, 2024.

[12] P. Li, "Performance Boost in Deep Neural Networks: Improved ResNext50 for Complex Image Datasets," Transactions on Computational and Scientific Methods, vol. 5, no. 1, 2025.

[13] X. Wang, "Mining Multimodal Data with Sparse Decomposition and Adaptive Weighting," Transactions on Computational and Scientific Methods, vol. 5, no. 1, 2025.

[14] S. Wang, C. Wang, J. Gao, Z. Qi, H. Zheng, and X. Liao, "Feature Alignment-Based Knowledge Distillation for Efficient Compression of Large Language Models," arXiv preprint, arXiv:2412.19449, 2024.

[15] X. Li, T. Ruan, Y. Li, Q. Lu, and X. Sun, "A Matrix Logic Approach to Efficient Frequent Itemset Discovery in Large Data Sets," arXiv preprint, arXiv:2412.19420, 2024.

[16] R. Zhang, S. Wang, T. Xie, S. Duan, and M. Chen, "Dynamic User Interface Generation for Enhanced Human-Computer Interaction Using Variational Autoencoders," arXiv preprint, arXiv:2412.14521, 2024.

[17] Y. Liang, E. Gao, Y. Ma, Q. Zhan, D. Sun, and X. Gu, "Contextual Analysis Using Deep Learning for Sensitive Information Detection," Proceedings of the 2024 International Conference on Computers, Information Processing and Advanced Education (CIPAE), pp. 633–637, 2024.

[18] P. Li, "Machine Learning Techniques for Pattern Recognition in High-Dimensional Data Mining," arXiv preprint, arXiv:2412.15593, 2024.

[19] C. Ruan, C. Huang, and Y. Yang, "Comprehensive Evaluation of Multimodal AI Models in Medical Imaging Diagnosis: From Data Augmentation to Preference-Based Comparison," arXiv preprint, arXiv:2412.05536, 2024.

[20] X. Yan, W. Wang, M. Xiao, Y. Li, and M. Gao, "Survival prediction across diverse cancer types using neural networks", Proceedings of the 2024 7th International Conference on Machine Vision and Applications, pp. 134-138, 2024.

[21] S. Khan, M. Naseer, M. Hayat, S. W. Zamir, F. S. Khan, and M. Shah, "Transformers in vision: A survey," ACM Computing Surveys, vol. 54, no. 10s, pp. 1-41, 2022.

[22] W. He, T. Zhou, Y. Xiang, Y. Lin, J. Hu, and R. Bao, "Deep Learning in Image Classification: Evaluating VGG19's Performance on Complex Visual Data," *arXiv preprint arXiv:2412.20345*, 2024.

[23] A. Priya and P. S. Bharathi, "SE-ResNeXt-50-CNN: A Deep Learning Model for Lung Cancer Classification," *Applied Soft Computing*, vol. 112696, 2025.